\documentclass[conference, a4paper]{IEEEtran}
\usepackage{capt-of}
\usepackage{microtype}
\usepackage{cite}
\usepackage{amsmath,amssymb,amsfonts}
\usepackage{graphicx}
\usepackage{textcomp}
\usepackage{xcolor}
\usepackage{hyperref}
\usepackage{enumitem}

\hypersetup{
    colorlinks=true,
    linkcolor=black,
    citecolor=blue,
    filecolor=black,
    urlcolor=black,
}

\begin{document}

\title{Beyond Sequences: A Benchmark for Atomic Hand-Object Interaction Using a Static RNN Encoder}

\author{
    \IEEEauthorblockN{Yousef Azizi Movahed}
    \IEEEauthorblockA{
        \textit{Dept. of Computer Science, School of Mathematics,} \\
        \textit{Statistics, and Computer Science, College of Science} \\
        \textit{University of Tehran}\\
        Tehran, Iran \\
        azizi.movahed@ut.ac.ir
    }
    \and
    \IEEEauthorblockN{Fatemeh Ziaeetabar}
    \IEEEauthorblockA{
        \textit{Dept. of Computer Science, School of Mathematics,} \\
        \textit{Statistics, and Computer Science, College of Science} \\
        \textit{University of Tehran}\\
        Tehran, Iran \\
        fziaeetabar@ut.ac.ir
    }
}

\maketitle

\begin{abstract}
Reliably predicting human intent in hand-object interactions is an open challenge for computer vision. Our research concentrates on a fundamental sub-problem: the fine-grained classification of atomic interaction states, namely 'approaching', 'grabbing', and 'holding'. To this end, we introduce a structured data engineering process that converts raw videos from the MANIAC dataset into 27,476 statistical-kinematic feature vectors. Each vector encapsulates relational and dynamic properties from a short temporal window of motion. Our initial hypothesis posited that sequential modeling would be critical, leading us to compare static classifiers (MLPs) against temporal models (RNNs). Counter-intuitively, the key discovery occurred when we set the sequence length of a Bidirectional RNN to one (seq\_length=1). This modification converted the network's function, compelling it to act as a high-capacity static feature encoder. This architectural change directly led to a significant accuracy improvement, culminating in a final score of 97.60\%. Of particular note, our optimized model successfully overcame the most challenging transitional class, 'grabbing', by achieving a balanced F1-score of 0.90. These findings provide a new benchmark for low-level hand-object interaction recognition using structured, interpretable features and lightweight architectures.
\end{abstract}


\section{Introduction}
\label{sec:intro}

Effective human-machine collaboration requires systems to move beyond recognizing current actions to anticipating future intent. This ability, "action anticipation," is foundational for robotics and HCI. However, anticipating complex activities (e.g., 'making a sandwich') first requires granular classification of the underlying atomic interaction states. These states (e.g., 'approaching', 'grabbing', 'holding') are the building blocks of all object manipulation tasks.

Recent advances in action understanding center on end-to-end models like 3D-CNNs and Vision Transformers~\cite{carreira2017i3d, arnab2021vivit}. While successful at high-level action recognition, their design differs from our task. They classify extended videos (e.g., 'Making a Sandwich') from raw pixels, whereas we identify short-term atomic states (e.g., 'grabbing') from structured features. These models also differ in modality (e.g., raw pixels vs. skeletal data~\cite{yan2018stgcn}) and cannot benchmark performance on structured kinematic representations. Consequently, the performance ceiling for classifying atomic states using structured features remains underexplored. Without this core benchmark, evaluating complex anticipatory models is difficult.

In this study, we address this gap by establishing a performance benchmark for classifying low-level interaction states using a rigorous data engineering pipeline. We transformed raw MANIAC videos~\cite{aksoy2014maniac} into a structured dataset of statistical–kinematic feature vectors. Our investigation led to three main contributions:

\begin{itemize}[leftmargin=*, noitemsep, nolistsep, topsep=2pt, partopsep=0pt]
   \item We provide a new benchmark for atomic HOI state classification using structured, statistical-kinematic features.
   \item We find that temporal context is non-essential, as our key breakthrough emerged from a Bidirectional RNN with \texttt{seq\_length=1} (a static encoder).
   \item This 'static RNN' approach achieves \textbf{97.60\%} accuracy and, crucially, resolves the most challenging transitional class, 'grabbing', with a balanced F1-score of \textbf{0.90}.
\end{itemize}

\section{Related Work}
\label{sec:related}

\subsection{Cognitive Foundations and Computational Models}
Our work on action anticipation—predicting future actions from early cues~\cite{kong2019action}—draws from cognitive science. Theories such as Gibson’s affordances~\cite{gibson1979ecological} and Predictive Coding~\cite{friston2005theory} imply that models must look beyond simple motion and incorporate agent-object relationships. The complexity of bimanual actions, as described by Guiard~\cite{guiard1987asymmetric}, further reinforces the inherently relational nature of these interactions. This perspective motivates using structured representations, such as Graph Neural Networks (GNNs)~\cite{scarselli2008graph} or Multi-Agent Systems (MAS)~\cite{wooldridge2009introduction}.

Computational action understanding has evolved from monolithic ”black-box” approaches to structured approaches. Early models combined CNNs with RNNs~\cite{donahue2015}, but these were soon superseded by end-to-end 3D-CNNs (e.g., I3D)~\cite{carreira2017i3d} and Video Transformers~\cite{arnab2021vivit}. Although powerful, these models are opaque and lack explicit representations of scene structures (agents, objects, and their relationships), a trait that hinders fine-grained relational reasoning. This limitation prompted a shift toward graph-based representations. This trend began in skeleton-based action recognition (e.g., ST-GCN~\cite{yan2018stgcn}), which demonstrated the power of graph-based reasoning. Concurrently, semantic frameworks have modeled manipulation from a relational perspective~\cite{ziaeetabar2017semantic, ziaeetabar2018prediction, ziaeetabar2018recognition}. A logical next step was incorporating objects into the graph, driving human–object interaction (HOI) research~\cite{qi2018hoi, gao2020drg}. However, most HOI studies focus on high-level detection (e.g., ”person opening door”) and not the fine-grained, split-second atomic states (e.g., grabbing) that we address. Recent work on bimanual manipulation using hierarchical graphs~\cite{ziaeetabar2024a, ziaeetabar2024b} confirms the value of structured modeling but does not establish a benchmark for these specific low-level tasks.

\subsection{Identifying the Research Gap}
A review of the literature, from early pixel-based models to modern structured approaches for bimanual actions, reveals a clear and critical gap. The field lacks a reliable, high-performance benchmark for the specific intermediate task of classifying atomic interaction states (such as ’grabbing’) from statistical-kinematic features. 

Current SOTA models for action recognition present an unsuitable comparison because they are designed to process raw pixels (like I3D) or full-body skeletons (like ST-GCN) for high-level, whole-action labels. They do not provide a baseline for our more granular task, nor do they operate in the same feature-engineered input space. We argue that accurately predicting complex multi-step goals is only feasible after this fundamental classification of atomic actions has been reliably solved. This study was undertaken to provide such a benchmark. In doing so, we also arrived at a counterintuitive conclusion regarding the optimal modeling approach for what appears to be a straightforward classification task.

\section{Methodology}
\label{sec:methodology}

\begin{figure*}[htbp]
\centering
\includegraphics[width=0.9\textwidth]{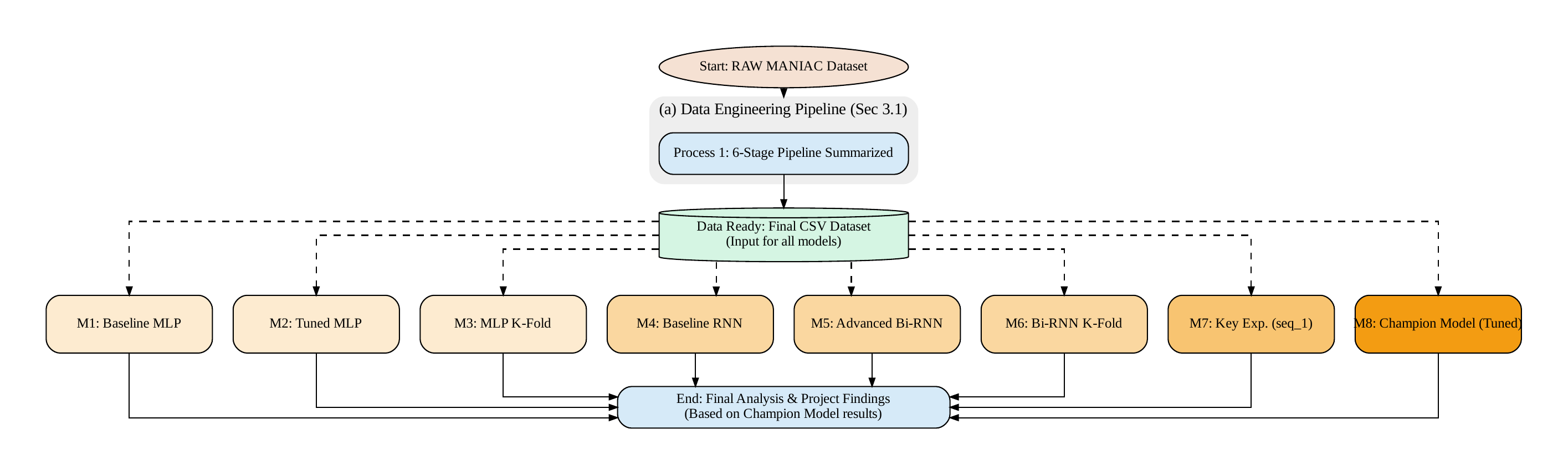}
\caption{Overview of the methodology. (a) The six-stage data-engineering pipeline converts raw MANIAC videos into a structured feature corpus. (b) The experimental evolution traces eight successive modeling stages, from static baselines to the final champion model.}
\label{fig:methodology_overview}
\end{figure*}

Our methodology consists of two tightly coupled parts: (1) construction of a structured dataset from raw video recordings, and (2) an iterative model development process that identifies and refines an effective inductive bias for the classification task. We first describe the data-engineering pipeline that transforms MANIAC videos into compact, information-rich descriptors; we then outline the sequence of modeling experiments that led to the final, optimized architecture. The full workflow is summarized in Fig.~\ref{fig:methodology_overview}.

\subsection{Data engineering and feature extraction}
\label{subsec:data_engineering}

Recognizing that input representation critically determines downstream performance for fine-grained interaction classification, we deliberately opted for feature engineering over end-to-end pixel learning. Using RGB frames and segmentation masks provided by the \textbf{MANIAC} dataset~\cite{aksoy2014maniac}, we implemented a deterministic pipeline that outputs fixed-length statistical–kinematic vectors. Across the corpus this procedure produced \(\mathbf{27,476}\) labeled examples, each summarizing the dynamics within a short temporal window.

\begin{figure}[htbp]
\centering
\includegraphics[width=0.9\columnwidth]{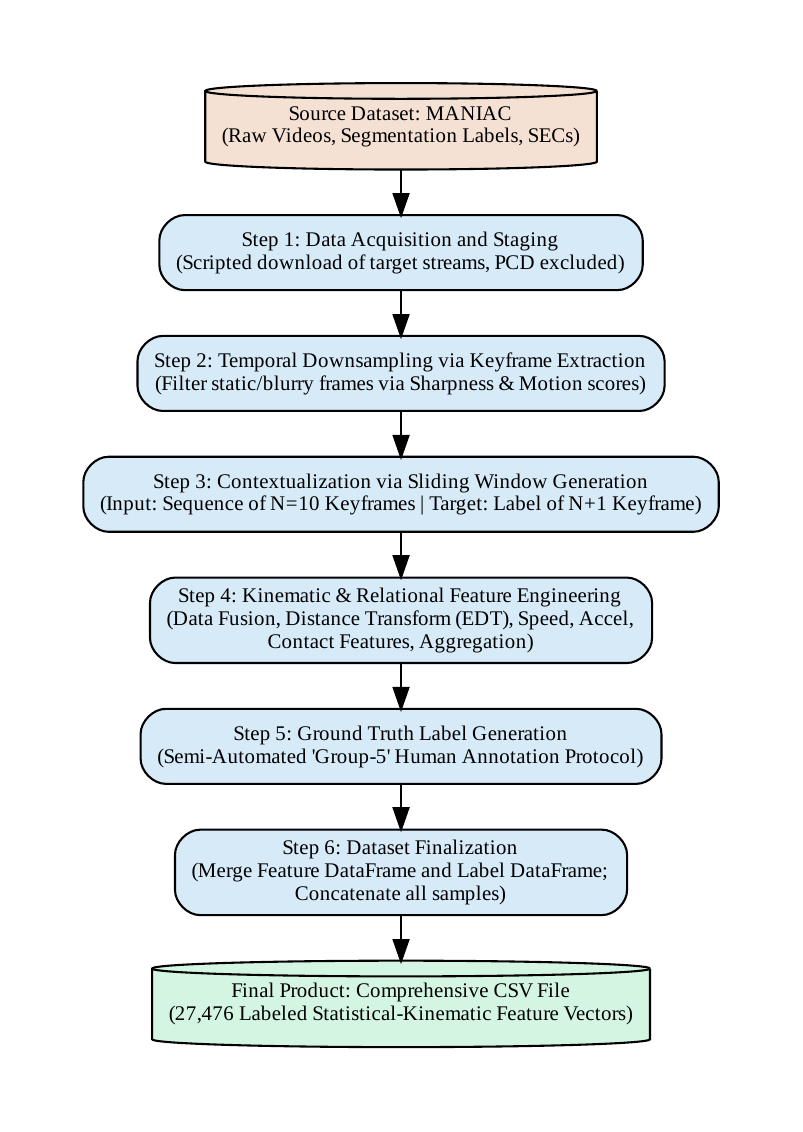}
\caption{The six-stage pipeline used to construct the statistical–kinematic dataset from MANIAC.}
\label{fig:dataset_pipeline_flowchart}
\end{figure}

Figure~\ref{fig:dataset_pipeline_flowchart} depicts the six principal stages of the pipeline.

The stages are as follows:

\textbf{1) Keyframe extraction (temporal downsampling).} To eliminate redundant frames while preserving salient motion cues, we selected keyframes using two concurrent criteria: image sharpness (quantified via Laplacian variance) and frame-difference energy. Frames surpassing predefined thresholds on both measures were retained, substantially reducing temporal redundancy and focusing subsequent processing on informative instants.

\begin{figure}[t]
\centering
\includegraphics[width=0.9\columnwidth]{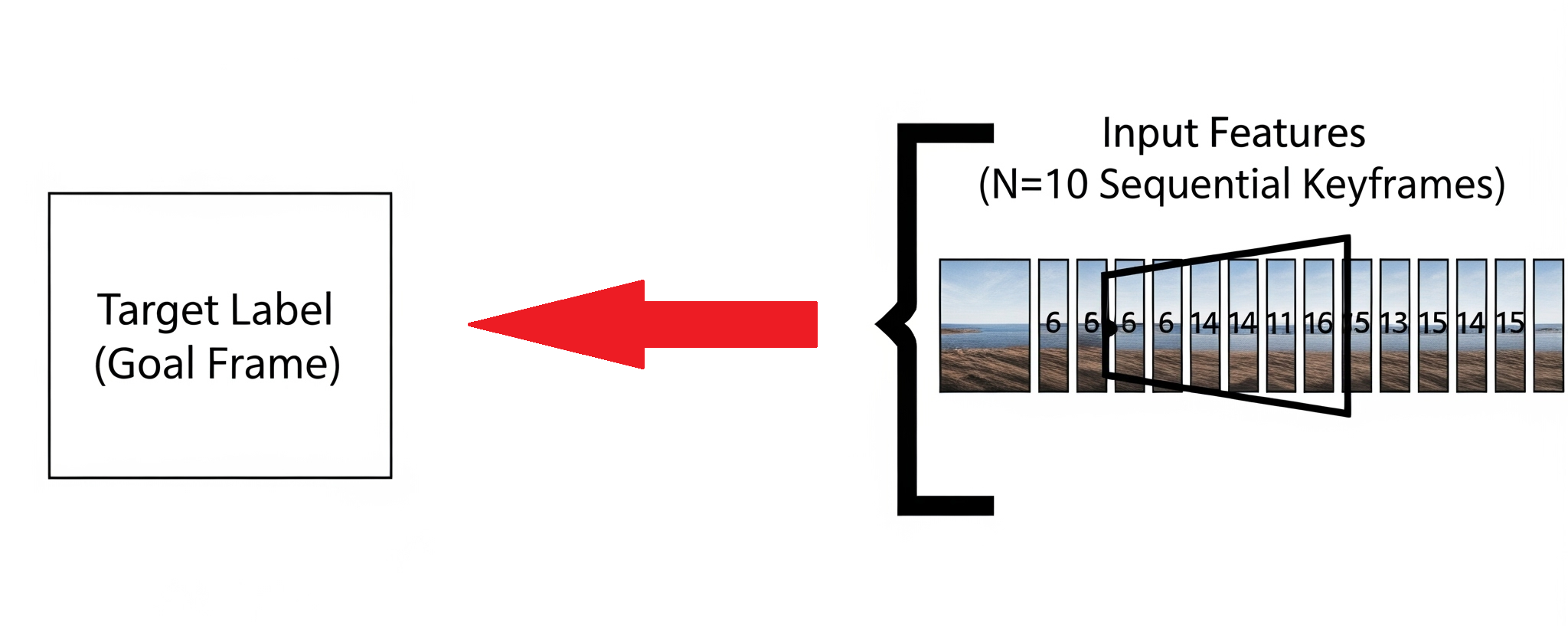}
\caption{Predictive sliding-window mechanism: a history of 10 keyframes is summarized into one statistical feature vector which is used to predict the label at the 11th keyframe.}
\label{fig:sliding_window}
\end{figure}

\textbf{2) Predictive windowing.} We framed the problem as anticipation by applying a sliding predictive window over the extracted keyframes. Concretely, a history of \(N=10\) keyframes serves as the input context, while the label corresponds to the \(11^{\mathrm{th}}\) keyframe (see Fig.~\ref{fig:sliding_window}). Each such window yields a single training sample.

\textbf{3) Kinematic and relational descriptors.} Each predictive window is converted to a compact feature vector by computing descriptors that capture spatial relations, motion dynamics, and contact phenomena.

\begin{itemize}
\item \textit{Robust proximity.} Instead of centroid-to-centroid distance, we compute an Euclidean Distance Transform (EDT) on the object mask to obtain a distance field; the minimal EDT value sampled within the hand mask provides a robust measure of hand–object proximity that is insensitive to complex object geometries.
\item \textit{Hand kinematics.} The hand centroid is tracked through the keyframes to derive per-frame velocities. From this sequence we extract summary statistics (e.g., mean speed and the standard deviation of speed) that effectively capture transient accelerations and decelerations associated with action transitions.
\item \textit{Contact-derived metrics.} Contact is operationalized via an EDT threshold ($\varepsilon = 10$ pixels). From the derived binary contact signal we compute \textit{Contact Count} (total frames in contact within the window) and \textit{Contact Duration} (maximum length of consecutive-contact frames), the latter distinguishing fleeting touches from sustained grasps.
\end{itemize}

All per-frame measurements (distances, velocities, contact flags) within a window are aggregated using standard statistical descriptors (mean, variance) alongside a simple linear-trend term, yielding a fixed-length representation per window. Repeating this procedure across the dataset produced the final set of 27,476 labeled feature vectors.

\subsection{Model design and experimental evolution}
\label{subsec:model_evolution}

\begin{figure}[htbp]
\centering
\includegraphics[width=\columnwidth]{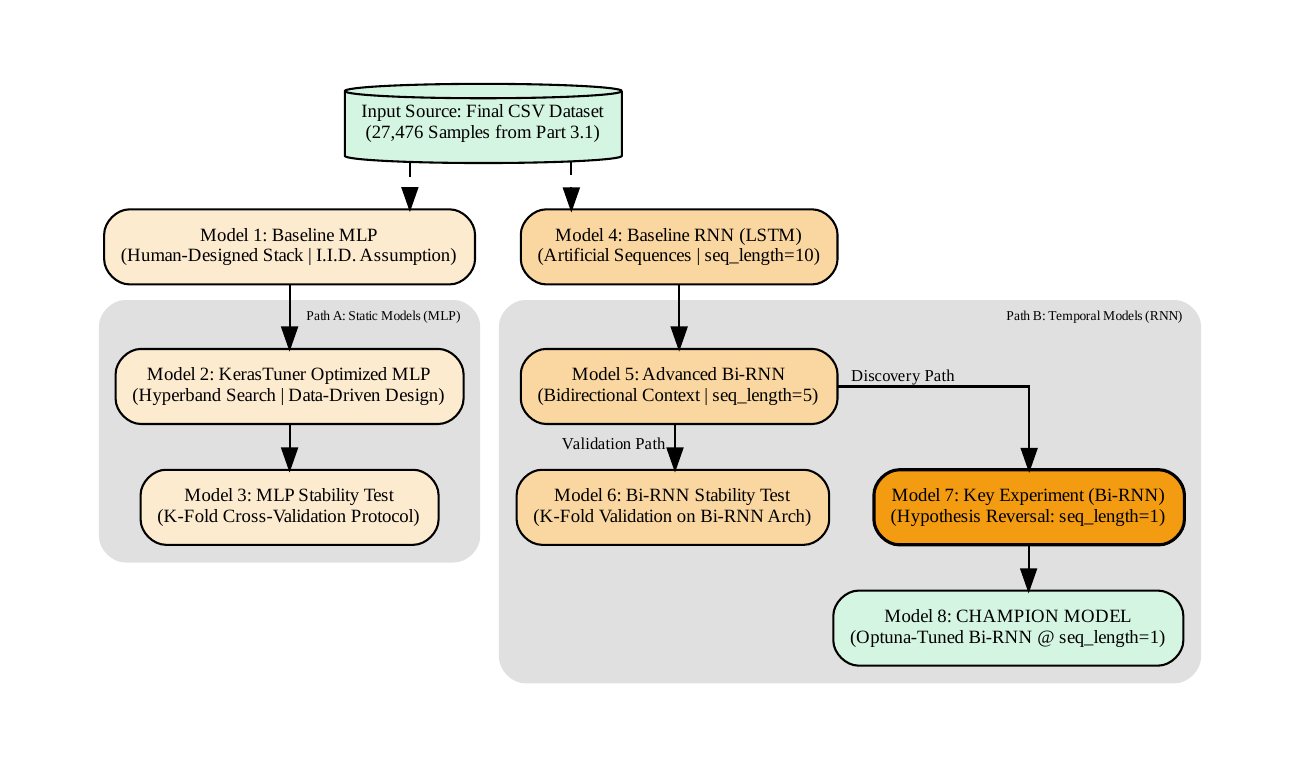}
\caption{Evolution of modeling experiments: from static MLPs through temporal RNNs to the final Optuna-optimized champion.}
\label{fig:models_evolution_flowchart}
\end{figure}

Model development proceeded as an empirical, hypothesis-driven exploration comprising eight experiments (summarized in Fig.~\ref{fig:models_evolution_flowchart}). Each stage tested a specific inductive assumption and informed subsequent design choices, moving from a conservative static baseline toward a strongly regularized, high-capacity encoder.

\subsubsection{Static baseline: optimized MLP}
We established a rigorous static baseline using a multilayer perceptron composed of dense layers (128–64–32), augmented with (L$_2$) regularization, batch normalization, and dropout. Class imbalance was managed via \texttt{class\_weight='balanced'}. Hyperparameters were tuned with \texttt{KerasTuner} to ensure the baseline represented a fair, well-optimized static comparator. Despite strong overall performance, the MLP struggled to disambiguate the transitional \textit{grabbing} state.

\subsubsection{Temporal modeling: bidirectional RNNs}
To evaluate whether explicit temporal context could resolve this weakness, we developed recurrent architectures, including LSTM and bidirectional RNN variants. When configured with multi-step sequences (for example, \texttt{seq\_length=5}), the bidirectional models captured bidirectional temporal dependencies and matched the MLP’s accuracy, but offered limited benefit for the problematic \textit{grabbing} class.

\subsubsection{Reframing the RNN as a static encoder}
Confronted with an empirical performance plateau across both static and temporal models, we hypothesized that the engineered feature vectors already encapsulate essential discriminative information and that additional temporal aggregation might introduce noise. To test this, we constrained the bidirectional RNN to a single time step (\texttt{seq\_length=1}). In this configuration the recurrent cell is not used for temporal unrolling; rather, its internal gates and nonlinear transformations function as a deep, expressive encoder that processes each feature vector independently.

This modification produced a marked and consistent improvement in classification performance, indicating that the representational power of a gated recurrent cell—applied as a static encoder—better captures the complex, high-dimensional relationships encoded by our aggregated statistical features than the previously used dense stacks.

\subsubsection{Champion model: Optuna-driven refinement}
\begin{figure}[htbp]
\centering
\includegraphics[width=1.0\columnwidth]{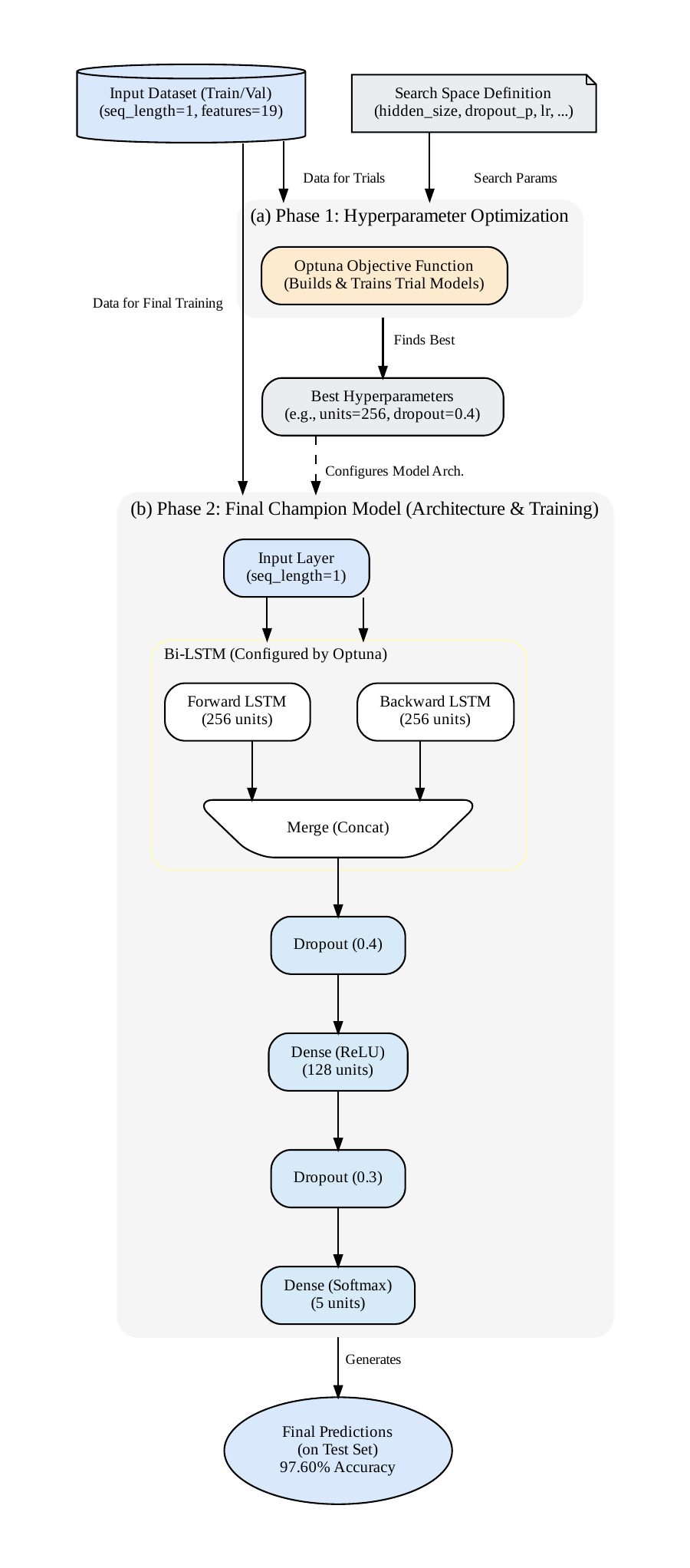}
\caption{Champion architecture: an Optuna-optimized bidirectional RNN configured with \texttt{seq\_length=1}, serving as a high-capacity static encoder for individual feature vectors.}
\label{fig:champion_model_arch}
\end{figure}

Having validated the \texttt{seq\_length=1} Bi-RNN as the most effective structural bias, we performed an extensive hyperparameter optimization using the Optuna framework. The search space included the number of units and layers, dropout probabilities, learning rate, and batch size. The resulting Optuna-tuned configuration constitutes our champion model (illustrated in Fig.~\ref{fig:champion_model_arch}) and forms the basis for the benchmark results reported in Section~\ref{sec:results}.

\section{Experiments and Results}
\label{sec:results}

\begin{table*}[!t]
\centering
\caption{Comprehensive Performance Comparison of All 8 Models Along Our Evolutionary Path}
\label{tab:results_summary_full}
\renewcommand{\arraystretch}{1.3} 
\small
\begin{tabular*}{\textwidth}{@{\extracolsep{\fill}}llccccc}
\hline\hline
\textbf{Model No.} & \textbf{Model Description} & \textbf{Architecture} & \textbf{Seq Len} & \textbf{Overall Acc.} & \textbf{Weighted F1} & \textbf{Grabbing F1-Score} \\
\hline
\multicolumn{7}{l}{\textit{Phase 1: Static MLP Baselines}} \\
Model 1 & Baseline MLP & MLP & N/A & 80.37\% & 0.82 & 0.43 (P:0.31, R:0.72) \\
Model 2 & KerasTuner-Optimized MLP & MLP & N/A & 89.45\% & 0.88 & 0.45 (P:0.84, R:0.31) \\
Model 3 & K-Fold Validation of MLP & MLP & N/A & 87.89\% (avg) & - & - \\
\hline
\multicolumn{7}{l}{\textit{Phase 2: The Temporal Hypothesis}} \\
Model 4 & Baseline RNN (Flawed) & LSTM & 10 & 82.91\% & 0.81 & 0.34 \\
Model 5 & Advanced Bi-RNN & Bi-RNN & 5 & 89.35\% & 0.89 & 0.55 \\
Model 6 & K-Fold Validation of Bi-RNN & Bi-RNN & 5 & 89.00\% (avg) & 0.89 (avg) & 0.60 (avg) \\
\hline
\multicolumn{7}{l}{\textit{Phase 3: The Breakthrough and Final Optimization}} \\
\textbf{Model 7} & \textbf{The Key Experiment (Static RNN)} & \textbf{Bi-RNN} & \textbf{1} & \textbf{97.33\%} & \textbf{0.97} & \textbf{0.89} \\
\textbf{Model 8} & \textbf{Champion Model (Optuna-Tuned)} & \textbf{Bi-RNN} & \textbf{1} & \textbf{97.60\%} & \textbf{0.98} & \textbf{0.90 (P:0.90, R:0.90)} \\
\hline\hline
\end{tabular*}
\end{table*}

To ensure full transparency and reproducibility, the complete benchmark dataset (27,476 feature vectors) and model code are publicly available in our project repository \cite{Azizi2025BeyondSequencesCode}. We trained all models on 80\% of the dataset and evaluated them on a held-out 20\% test set. The results of our eight-stage investigation are summarized in Table~\ref{tab:results_summary_full}.

\subsection{Initial Performance Plateau}
Our initial experiments established a clear performance ceiling, as detailed in Table~\ref{tab:results_summary_full}. Both the optimized static classifier (Model 2, MLP) and the advanced temporal model (Model 6, Bi-RNN w/ \texttt{seq\_length=5}) plateaued at \(\approx 89\%\) accuracy. A key shortcoming was evident in both architectures: neither could resolve the ambiguous \texttt{grabbing} class, whose F1-scores remained persistently low. This consistent failure across models compelled us to re-evaluate our initial temporal hypothesis.

\subsection{The Breakthrough: An RNN as a Static Encoder}
The stagnation at the \(\approx 89\%\) accuracy threshold motivated a pivotal experiment (Model 7). We reversed our original hypothesis, positing that the temporal context \textit{between} feature vectors was not only unhelpful but actively introduced noise. We theorized that the essential information was already fully embedded \textit{within} each individual vector.

\begin{figure}[!ht]
\centering
\includegraphics[width=1.0\columnwidth]{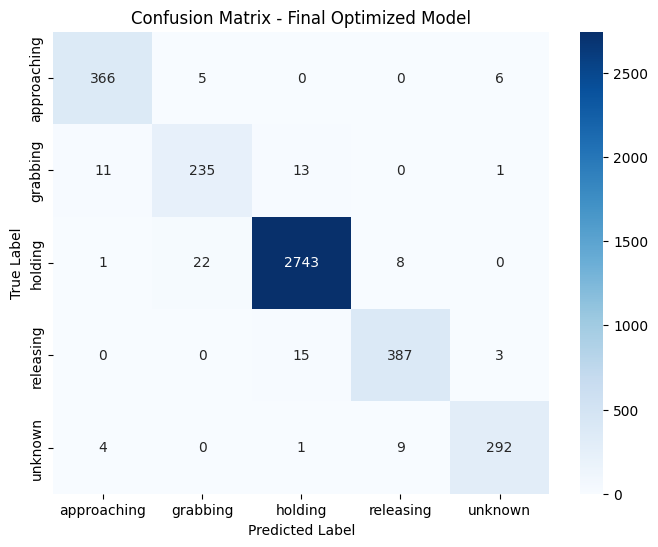}
\caption{Confusion Matrix for the Champion Model (Model 8). The confusion matrix shows minimal off-diagonal errors, indicating excellent class separation.}
\label{fig:champion_confusion_matrix}
\end{figure}

To test this, we configured the Bi-RNN with \texttt{seq\_length=1}. As shown in Table~\ref{tab:results_summary_full}, the impact was immediate: overall accuracy surged to 97.33\%, while the F1-score for the previously challenging \texttt{grabbing} class rose to 0.89. \textbf{This result was significant, as the 'grabbing' class is not only a brief transitional state but also the most under-represented in the dataset (see Fig.~\ref{fig:champion_confusion_matrix}).}

This outcome provides strong empirical support for our revised hypothesis. We posit that the internal gating mechanisms of an RNN cell, even when unrolled for a single time step, offer a significantly more expressive and nonlinear feature transformation compared to a conventional stack of dense layers.

In this configuration, setting \texttt{seq\_length=1} is no longer a temporal analysis; rather, we repurposed the sophisticated RNN architecture to function as a high-capacity static feature encoder. This encoder proved exceptionally effective at modeling the intricate, high-dimensional relationships embedded within our statistical vectors—relationships that the simpler MLP (Model 2) failed to capture.

Our final step was to refine this new paradigm. We employed the Optuna framework to exhaustively optimize the \texttt{seq\_length=1} Bi-RNN, resulting in our Champion Model (Model 8). This model achieved a final accuracy of \textbf{97.60\%}. As shown in Fig.~\ref{fig:champion_confusion_matrix} and Table~\ref{tab:champion_class_report}, the model's performance demonstrates near-perfect classification. The confusion matrix is almost perfectly diagonal, and the model achieves a balanced F1-score of \textbf{0.90} for the 'grabbing' class. This work therefore resolves the core classification challenge and establishes a new, robust performance benchmark for this foundational task.

\begin{table}[!t]
\centering
\caption{Classification Report for the Champion Model (Model 8)}
\label{tab:champion_class_report}
\renewcommand{\arraystretch}{1.25}
\resizebox{\columnwidth}{!}{%
\begin{tabular}{l c c c r}
    \hline\hline
    \multicolumn{5}{c}{\textbf{Classification Report:}} \\
    \hline
    & \textbf{precision} & \textbf{recall} & \textbf{f1-score} & \textbf{support} \\
    \hline
    approaching & 0.96 & 0.97 & 0.96 & 377 \\
    grabbing    & 0.90 & 0.90 & 0.90 & 260 \\
    holding     & 0.99 & 0.99 & 0.99 & 2774 \\
    releasing   & 0.96 & 0.96 & 0.96 & 405 \\
    unknown     & 0.97 & 0.95 & 0.96 & 306 \\
    \hline
    accuracy    & \multicolumn{3}{c}{0.98} & 4122 \\
    macro avg   & 0.95 & 0.95 & 0.95 & 4122 \\
    weighted avg & 0.98 & 0.98 & 0.98 & 4122 \\
    \hline\hline
\end{tabular}%
}
\end{table}

\subsubsection*{Clarification of Atomic States}
For clarity, the core interaction states classified in this benchmark are operationally defined as:
\begin{itemize}
    \item \textbf{approaching:} The hand is in motion towards the object to grab it.
    \item \textbf{grabbing:} The brief transitional state, on the verge of contact, where the hand is ready to secure the object.
    \item \textbf{holding:} The object is fully secured and stable within the hand.
    \item \textbf{releasing:} The hand is opening, having just let go of the object.
    \item \textbf{unknown:} Any state not matching the above criteria (e.g., hand is static and far from the object).
\end{itemize}

\section{Conclusion}
\label{sec:conclusion}

The classification of atomic manipulation states in human-object interaction has lacked a robust benchmark. In this study, we addressed this gap by first engineering a rich dataset of statistical–kinematic features, derived from a careful data pipeline, to capture fine-grained dynamics. The investigation culminated in a model achieving 97.60\% accuracy, setting a new baseline by successfully resolving ambiguous transitional states.

The 97.60\% accuracy, while significant, is not the central contribution. The key finding is counter-intuitive: our initial hypothesis, which assumed temporal context between windows was essential, was incorrect. The results clearly show that for these richly engineered features, the static encoder's representational power is far more critical than the temporal relationships between samples.

It is important to situate this work correctly. While Temporal Transformers excel in sequence modeling, our goal was not end-to-end learning. The focus was on high-fidelity classification from pre-engineered features. This strategy reduced both data volume and dimensionality, allowing temporal analysis to be applied to compact representations, not full sequences.

Within this defined scope, established gated architectures (RNNs and LSTMs) proved highly effective. They offered a solution that was not only robust but also efficient in terms of computational load and hardware demands, side-stepping the implementation overhead of newer models. This balance represented a pragmatic choice for the task.

This is not to dismiss the potential of modern architectures. The benchmark established here provides a solid foundation for a transition to dynamic prediction. Future work will naturally explore advanced models (like GNNs) for complex multi-agent interactions, and our current classifier will serve as a firm baseline for evaluating those systems.

To solidify these findings, three avenues remain critical: (1) validating the pipeline and the "static RNN" concept on other HOI datasets (e.g., H2O, EGTEA); (2) using interpretability methods (like SHAP) to dissect \textit{how} the RNN's gating mechanism achieves a superior representation from our features compared to the baseline MLP; and (3) progressing from manual feature engineering towards end-to-end feature learning, particularly with grounded visual inputs.

\section*{Acknowledgment}

\bibliographystyle{IEEEtran}
\bibliography{references}

\end{document}